\documentclass[letterpaper]{article} 
\usepackage[]{aaai23}  
\usepackage{times}  
\usepackage{helvet}  
\usepackage{courier}  
\usepackage[hyphens]{url}  
\usepackage{graphicx} 
\usepackage{natbib}  
\usepackage{caption} 
\usepackage{algorithm}
\usepackage{algorithmic}

\usepackage{tikz}
\usetikzlibrary{shapes.geometric, arrows}
\tikzstyle{startstop} = [rectangle, rounded corners, minimum width=0.7cm, minimum height=0.7cm,text centered, draw=black, fill=red!20]

\tikzstyle{judge1} = [rectangle, minimum width=1cm, minimum height=0.7cm, text centered,  draw=black, fill=blue!20]
\tikzstyle{process1} = [rectangle, minimum width=1cm, minimum height=0.7cm, text centered,  draw=black, fill=blue!20]
\tikzstyle{decision1} = [rectangle, minimum width=1cm, minimum height=0.7cm, text centered, draw=black, fill=blue!20]

\tikzstyle{judgedots} = [rectangle, minimum width=1cm, minimum height=0.7cm, text centered,  draw=black, fill=orange!20]
\tikzstyle{processdots} = [rectangle, minimum width=1cm, minimum height=0.7cm, text centered,  draw=black, fill=orange!20]
\tikzstyle{decisiondots} = [rectangle, minimum width=1cm, minimum height=0.7cm, text centered, draw=black, fill=orange!20]

\tikzstyle{judge9} = [rectangle, minimum width=1cm, minimum height=0.7cm, text centered,  draw=black, fill=green!20]
\tikzstyle{process9} = [rectangle, minimum width=1cm, minimum height=0.7cm, text centered,  draw=black, fill=green!20]
\tikzstyle{decision9} = [rectangle, minimum width=1cm, minimum height=0.7cm, text centered, draw=black, fill=green!20]

\tikzstyle{arrow} = [thick,->,>=stealth]

\usepackage{newfloat}
\usepackage{listings}
\DeclareCaptionStyle{ruled}{labelfont=normalfont,labelsep=colon,strut=off} 
\floatstyle{ruled}
\newfloat{listing}{tb}{lst}{}
\floatname{listing}{Listing}

\title{Blind Judgement: \\Agent-Based Supreme Court Modelling With GPT}
\author {
    Sil Hamilton
}
\affiliations{
    McGill University\\
    sil.hamilton@mcgill.ca
}

\begin{document}

\maketitle

\begin{abstract}
We present a novel Transformer-based multi-agent system for simulating the judicial rulings of the 2010-2016 Supreme Court of the United States. We train nine separate models with the respective authored opinions of each supreme justice active ca. 2015 and test the resulting system on 96 real-world cases. We find our system predicts the decisions of the real-world Supreme Court with better-than-random accuracy. We further find a correlation between model accuracy with respect to individual justices and their alignment between legal conservatism \& liberalism. Our methods and results hold significance for researchers interested in using language models to simulate politically-charged discourse between multiple agents.
\end{abstract}

\section{Introduction}
Recent and ongoing political turmoil in the United States has magnified the actions of the federal Supreme Court in the public eye. The Court has taken to overturning judicial precedent in recent years, with the number of such decisions in the last six years reaching over twice the number of overturns between 2010 to 2015\footnote{2010-2015: 8 overturns, 2016-2022: 22+ overturns.}. The weakening rule of \textit{stare decisis} has encouraged judicial researchers to develop holistic models of Supreme Court behaviour to better predict and account for future trends \cite{blake2019don,allcorn2021unpacking}.

Accurate models of Supreme Court behaviour are rare despite this focus. The best performing models only reach accuracy levels of $\approx 70\%$ on out-of-distribution cases \cite{10.1371/journal.pone.0174698}. Models achieving even this middling accuracy are complex in their architecture, generally consisting of a mix of SVM and logistic regression models. This complexity is necessitated by the variables involved.

Confounding variables discussed in the literature include little agreed-upon theories regarding the legal doctrines practiced by individual justices \cite{jrDecisionMakingModern2011} and their rarely-documented social realities \cite{kromphardt2017evaluating,peterson2021personal}. While the in-court behaviour of the justices is well documented, exogenous factors have an equal impact on case decision-making. A model capable of both cognitive and social reasoning would therefore benefit justice behaviour modelling. To this end, we investigate whether recent advances in social simulation with language models can promote simple and effective models of Supreme Court behaviour.

\section{Background}
In this section we describe the rationale behind our project.

\subsection{Judicial Modelling}
Three major theories of judicial behaviour generally inform the design of Supreme Court models: the legal theory, the attitudinal theory, and the strategic theory \cite{jrDecisionMakingModern2011}. The \emph{legal theory} suggests justices are bound by constitutional precedent. The \emph{attitudinal theory} instead argues justices account for policy preference first, precedent second. Between the two lies the \emph{strategic theory}, which says justices vote according to a mix of precedent and preference. 

\begin{figure}
  \centering
  \includegraphics[width=1\linewidth]{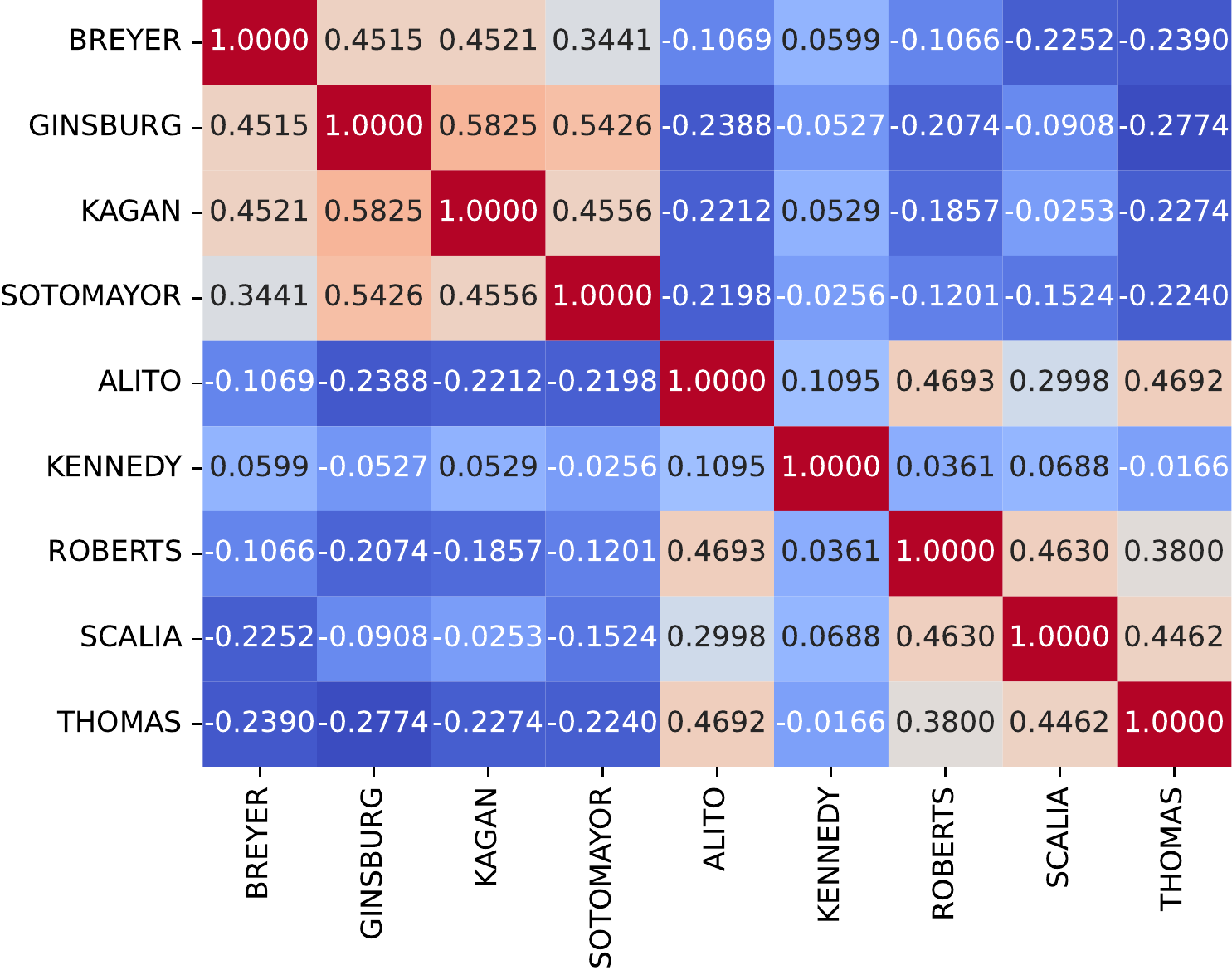}
  \caption{Correlation matrix of justices voting on 290 cases between 2010 and 2016. Note the clustering of justices nominated by Democrat and Republican presidents.}
  \label{figure:correlations}
\end{figure}

As we show in Figure 1, decision correlations between justices active between 2010 and 2016 (hereafter referred to as the Roberts IV court) indicate the strategic theory is most accurate to reality. While justices will invoke precedent when writing their rationales, evidence suggests justices remain somewhat beholden to the political alignment of their nominator. We note, however, that the correlations are only medium in their strength. This indicates accurate models should account for precedent, but not exclusively.

Integrating one of these three theories into a Supreme Court model requires choosing how to best cast the influence of precedence and preference as variables. Given this conversion can itself result in significant drawbacks via unforeseen factors, we instead choose a simulative tool which allows to us to make fewer assumptions as to the most correct theory of judicial behaviour: language models.

\subsection{Simulation}
Large Language Models (LLMs) are adept at simulating complex social phenomena. Recent research has demonstrated their ability to predict populated social media platforms \cite{https://doi.org/10.48550/arxiv.2208.04024}, the distribution of votes for presidential candidates in the 2012-2020 American elections \cite{argyle2022out}, and the general sentiment of news articles reporting COVID-19 in the early stages of the pandemic \cite{hamiltonCOVIDThatWasn2022}. These developments show model bias is valuable for those in the social sciences given bias is derived from the underlying distributions of their training material.

Prior simulation research benefits from new techniques for eliciting cognitive activity from LLMs nominally designed for next-token prediction. These include chain of thought reasoning \cite{wei2022chain}, discretely-structured prompts \cite{liu2021pre}, and fine-tuning \cite{drori2022neural}. These techniques have the model draw on internal biases to make predictions, allowing researchers to embed fewer assumptions into their simulative models. LLMs are alluring given judicial modelling necessitates a system capable of both social and cognitive reasoning.

\subsection{Agent-Based Modelling}
The process by which the Court arrives at a decision is nominally rational \cite{jrDecisionMakingModern2011}. While predilections are known to influence vote outcomes, justices are expected to justify dissenting decisions in written documents called \emph{opinions}.  Opinions are typically one to five pages in length within which the justice (or their aid) lays out their argument in a manner similar to an essay. For our simulation task, we assume justices record their rationale honestly and so treat the opinions as our primary target of prediction, meaning any model we train will be predicting opinions.

Because opinions are long documents (i.e. longer than the 1024 token-long context window GPT-2 is trained for), having one model produce multiple opinions in the same run is untenable. We turn to agent-based modelling for a solution.

Whether consolidating multiple generative LLMs into a single architecture is beneficial has been heretofore understudied, with the only significant prior experiment with LLMs showing promise \cite{betzNaturalLanguageMultiAgentSimulations2022}. However, given the success of the Mixture of Experts (MoE) method in machine translation \cite{https://doi.org/10.48550/arxiv.2207.04672}, we argue further exploration of similar techniques for simulation tasks is warranted. While social simulation experiments suggest a single language model is capable of producing a wide range of opinions, training multiple models separately prevents cross-contamination when studying multiple data sources.

\section{Method}
We present the general design of our architecture in three parts: data collection, system architecture, and measures.

\subsection{Dataset}
We source data from two datasets for this experiment. The first corpus is the Supreme Court Database (SCDB) released by researchers at Washington University in St. Louis, which provides variables for 9,095 cases decided between 1946 and 2021 \cite{spaeth2014supreme}. 

We supplement the SCDB with all written opinions from all slips provided on the Supreme Court website.\footnote{Found at https://www.supremecourt.gov/opinions/slipopinion} Extracting the opinions from the PDF documents with an optical character recognition (OCR) utility leaves us with 145MiB of text written between 2003 and 2022. We then associate each opinion with the justice and case from which it originated.

\subsection{System Architecture}
We choose to simulate the Roberts IV court (2010-2016) given this period outlasts all other Supreme Court iterations in recent history.\footnote{See http://scdb.wustl.edu/documentation.php?var=naturalCourt}

\begin{figure}
 \centering
\begin{tikzpicture}[node distance=1.8cm]

\node (case) [startstop] {Case Syllabus};

\node (judgedots) [judgedots, below of=case] {\dots};
\node (judge1) [judge1, left of=judgedots] {Judge 1};
\node (judge9) [judge9, right of=judgedots] {Judge $n$};

\node (opiniondots) [processdots, below of=judgedots] {\dots};
\node (opinion1) [process1, left of=opiniondots] {Opinion};
\node (opinion9) [process9, right of=opiniondots] {Opinion};

\node (decisiondots) [decisiondots, below of=opiniondots] {\dots};
\node (decision1) [decision1, left of=decisiondots] {Vote};
\node (decision9) [decision9, right of=decisiondots] {Vote};

\node (stop) [startstop, below of=decisiondots] {Majority Opinion};

\draw [arrow] (case) -- (judgedots);
\draw [arrow] (case) -- (judge1);
\draw [arrow] (case) -- (judge9);

\draw [arrow] (judgedots) -- (opiniondots);
\draw [arrow] (judge1) -- (opinion1);
\draw [arrow] (judge9) -- (opinion9);

\draw [arrow] (opiniondots) -- (decisiondots);
\draw [arrow] (opinion1) -- (decision1);
\draw [arrow] (opinion9) -- (decision9);

\draw [arrow] (decisiondots) -- (stop);
\draw [arrow] (decision1) -- (stop);
\draw [arrow] (decision9) -- (stop);

\end{tikzpicture}
\caption{Flow of our multi-agent system.}
\end{figure}
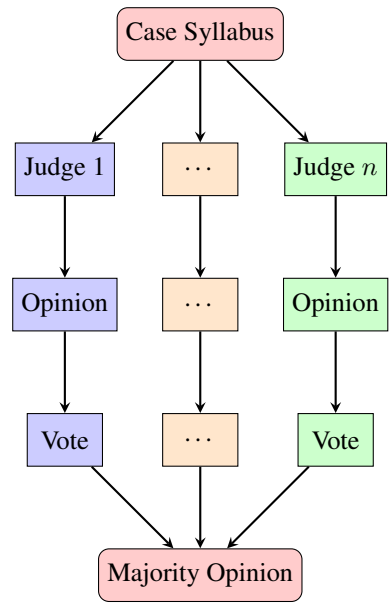

\paragraph{Design}
Our multi-agent system is composed of nine full-sized GPT-2 models \cite{radford2019language}. We present the system architecture in Figure 2. At a high level, our system receives the topic of a case being brought before the court and passes it along to nine justice models. The system then receives back nine opinions and corresponding decisions of whether to approve the appellant. The system totals the results and returns the majority vote.\footnote{We provide our code at [withheld from review copy]}

\paragraph{Prompt}
We train each justice model with a discrete prompt structured like a Python dictionary:

\begin{verbatim}
{
 'issue': 'Lorem ipsum…',
 'topic': 'Lorem ipsum…',
 'opinion': 'Lorem ipsum…'
 'decision': 'Lorem ipsum…'
}
\end{verbatim}

The \emph{issue} value corresponds to the \emph{issueArea} variable provided by SCDB.\footnote{See http://scdb.wustl.edu/documentation.php?var=issueArea} The \emph{topic} value is a short description of what the appellant is bringing before the court. We extract this information from the syllabus of each opinion slip and summarize it with GPT-3 Davinci \cite{brown2020language}. The \emph{opinion} value is the corresponding rationale the justice produces when formulating their \emph{decision} value, here a categorical variable signalling (dis)approval. We provide an example in the appendix.

\paragraph{Training} 
All models are trained for a total 30 epochs at a learning rate of 2$e^{-4}$ with the Adam optimizer \cite{https://doi.org/10.48550/arxiv.1412.6980}. This training process is conducted in two steps:
\begin{enumerate}
\item Construct $\leq1000$ token prompts of the above style for all cases in which the Roberts IV court came to a unanimous decision. This model serves as the base for all further trained models.
\item Collect all prompts generated in step 1 for each of the opinions (2003-2016) written by each justice active during Roberts IV. We thereby collect nine training sets and further train the model generated in step 1 with each separately.
\end{enumerate}
Average model loss after both steps is $1.5$, indicating there remains significant room for improvement.

\subsection{Measures}
We assess the performance of our multi-agent system on 96 test cases withheld from the training set with two measures: accuracy and a novel measure for judicial ideological alignment.

\paragraph{Accuracy} We measure accuracy with a receiver's operating characteristic curve (ROC) together with Cohen's $\kappa$ to account for a slight distribution bias in our test set.

\paragraph{Alignment} Justices are understood as being more or less in favour of overturning precedent. We capture this alignment by taking the Pearson coefficient ($r$) between model accuracy and the frequency with which the respective justice voted against precedent-altering decisions between 2003 and 2016. Our measure is intended to capture where a justice is aligned between conservative (e.g. textualism, formalism, originalism) or liberal (e.g. legal realism) frameworks of judicial decision making \cite{post2006originalism}.

\section{Results}
All results are reported with a minimum confidence rate of 80\% and are controlled for training material size and topic. Generations are run with a temperature of $0.5$ and a maximum length of 1000 tokens. 

\subsection{Accuracy}
Our system achieves an aggregate accuracy of 60\% ($\kappa \approx 0.18$) on 96 test cases. While less predictive than the state of the art, our model nonetheless achieves better-than-random performance despite having been trained solely on opinions.

{\renewcommand{\arraystretch}{1.2}
\begin{table}
\centering   
\begin{tabular}{| p{0.4\linewidth} c c |}
\hline
\textbf{Justice} & \textbf{Accuracy} & {$\kappa$} \\
\hline       
Samuel Alito & 65\% & 0.30 \\
Ruth Bader Ginsburg & 62\% & 0.21 \\
Clarence Thomas & 59\% & 0.18 \\
Stephen Breyer & 58\% & 0.16 \\
John Roberts & 57\% & 0.13 \\
Elena Kagan & 56\% & 0.12 \\
Anthony Kennedy & 54\% & 0.09 \\
Sonia Sotomayor & 51\% & 0.00 \\
Antonin Scalia & 50\% & -0.03 \\
\hline
\end{tabular} 
\caption{Model accuracy by justice. Note the wide variation in accuracy between justices.}
\end{table}}

We find a wide variation in the accuracy of each simulated justice when examining system performance more closely. As shown in Table 1, model accuracy varies between $65\%$ and $50\%$ despite having controlled for training data volume and case outcome.

\subsection{Alignment}
We  measure a moderate correlation ($r\approx0.56$) between simulated justice accuracy and the frequency with which each respective justice did not agree with the Court overruling or re-interpreting precedent. This result suggests our system achieves better accuracy with justices who are less likely to overturn precedent. We discuss the implications of this result below.

\subsection{Validation}
We train a single agent model to ensure having many agents provides non-negligible benefits. We fine-tune this single agent with the majority opinions of all cases decided on by the Roberts IV court. Testing this single agent on the test set results in an overall accuracy of 54\% ($\kappa = 0.08$). The predicted decisions differs from the original test set with a Cohen's $d$ of $\approx -0.86$ versus $d \approx 0.19$ for our multi-agent model, increasing the population overlap from 68.5\% to 92.4\%.

We implement software controls to ensure program output validity given training to a low loss does not guarantee the model produces both the \emph{opinion} and \emph{decision} variables. We therefore rerun each case until all models have returned a valid result. Once having processed all 96 cases, we sample agent-produced opinions belonging to half to ensure coherency.

\section{Discussion}
In this section we discuss two major consequences of our research.

\paragraph{Precedent Hallucination} GPT-2 is not an expert on legal precedent, nor should one expect it to be when the only formal source of legal information ingested by the model during pre-training were some seven thousand pages from \emph{FindLaw}, a website principally known for tort law \cite{clark_gpt-2_2022}.\footnote{https://www.findlaw.com/} This becomes evident when surveying model output. While the model will occasionally reference real laws, these citations prove to be happenstance as GPT-2 will confuse details and thus render the references meaningless.

That the model generates its own precedent when arguing over a case is an example of \emph{hallucination}, a well known property of language models \cite{rohrbach-etal-2018-object}. Because causal language models are only tasked with predicting the next most likely token given some prior sequence,  they are not given incentive to withhold factually incorrect statements---the model will say whatever is necessary to return the number of tokens requested in a cogent manner.

Our justice models will \emph{hallucinate precedence} when producing opinions. They produce this pretend precedence implicitly by citing it throughout the argumentation process. That our models achieve greater-than-random decision accuracy in voting outcomes despite not producing legally valid arguments suggests Supreme Court decisions may not always rest on legally coherent rationales.

\paragraph{Alignment Correlation} The correlation between model accuracy and judicial alignment indicates conservative justices are more predictable given their general unwillingness to overturn precedent. Considering the model hallucinates precedent, this correlation suggests conservative justices are conservative for ideological rather than rational reasons. 

We find this result surprising given conservative justices often make it a point to rationalize their unwillingness to overturn precedent with legal justifications. Common formalist theories of this sort include both originalism and textualism, doctrines practiced by conservative members of the current court \cite{esbeck2011uses}. Our results suggest these decision-making patterns are less grounded in rational logic than anticipated given they are partially captured in a model not familiar with common law.  

\section{Conclusion}
The aim of our project was to produce a multi-agent system capable of predicting Supreme Court decision-making with  little to no prior theory-based assumptions of judicial behaviour. Given our resulting model achieves better-than-random accuracy despite having been trained only on opinion matter, we argue our process serves as an example for researchers seeking to develop simulative experiments with language models.

\subsection{Limitations}
As should be expected of any project promoting the creative output of AI, we make note of the biased material used in the production of large language models like GPT-2. While we contend this culturally-derived bias is beneficial for researchers using foundation models in the the social sciences, we nonetheless ensure our model does not cause unwanted harm. As such, we clearly mark all samples as having been generated and refrain from releasing large collections of generated material to the public.

\subsection{Next Steps}
We propose the following next steps after having demonstrated the basic viability of our architecture.

\paragraph{Larger Model} Can we improve system accuracy with larger language models? Recent research suggests language models develop emergent cognitive features when scaled above 6.7 billion parameters, narrowing future possible candidates to the likes of GPT-NeoX-20B and T5X \cite{https://doi.org/10.48550/arxiv.2208.07339,GPT-NeoX-20B,roberts2022t5x}.

\paragraph{Larger Training Corpus} Another avenue for increasing system accuracy involves fine-training GPT-2 with the whole corpus of American law as captured by proceedings and opinions written in lower courts. The principle of \emph{stare decisis} means the practice of common law is a social venture, suggesting language models would do well in predicting precedent-dependent cases if prepared.

\paragraph{Improved Prompting} Research indicates language models can avoid the long tail of token probabilities by repetitively querying the model \cite{portelli2022generalizing, kim2020long}. Integrating repetitive prompting strategies into the opinion-generating schema is a promising avenue for improvement. Another avenue would be to assess how reinforcement learning from human feedback (RLHF) models like InstructGPT simulate court proceedings \cite{rlhf}.

\paragraph{Investigating Future Cases} How would the Roberts IV court fare with cases brought before the court after 2016? Would their court overturn precedent at the rate the post-2016 Supreme Court has? Questions of this caliber would be made approachable with a more accurate Roberts IV system.

\section{Acknowledgements}
I thank Prof. Andrew Piper of McGill University and Prof. Kristen Thomasen of UBC for their invaluable advice during the research process. I furthermore thank the reviewers and workshop committee members for their recommendations.

\bibliography{references.bib}

\appendix
\section{Appendix}
We demonstrate our system with a real-world case fed to the Ruth Bader Ginsburg model. We first provide our prompt followed by the model's response. Note the model correctly predicts Ginsburg's real-world decision (to deny the appellant) despite the "California Childhood Video Game Act" not existing. 

\subsection{Prompt}
\paragraph{Issue:} `First Amendment',
\paragraph{Topic:} `The California law restricting the sale or rental of violent video games to minors was challenged by the video-game and software industries. The Federal District Court concluded that the Act violated the First Amendment and permanently enjoined its enforcement. The Ninth Circuit affirmed.',
\paragraph{Appellant is seeking a:} `certiorari',

\subsection{Model Response}
\paragraph{Opinion:} `The California Childhood Video Game Act is on its face both overbroad and vague. It prohibits the sale or rental of “any device or medium designed for or capable of reproducing visual images of decapitation, blood, or dismemberment of a human body,” in the presence of a child under the age of 13. It also prohibits the importation into the State of California of “any video game system, computer game software, or computer program that contains a killing or violent content.” The Act specifically identifies video games that have been determined by the State to be “violent,” and any person who “possesses, or has under his control,” one of those video games is subject to the Act’s restrictions. The Act falls outside the First Amendment because it interferes with the creative expression and acquisitive impulse that go to make up the quintessential activity of a child’s life.

The majority points out that the Act does not regulate the mere possession of these games, but seeks to regulate the medium in which they are played. It reasons that the State may regulate what the child sees and does not see by preventing the child from becoming engrossed in the game world and thereby “sustaining or supporting the action or activity.” But the majority does not ask whether the regulation of what the child sees and does not see is content-based. It neither considers nor concludes whether the regulation is narrowly tailored to the interests it seeks to protect. And it does not consider the adverse impact that its rule may have on legitimate play. The majority says that the regulation is content-neutral because it does not regulate “the medium in which the [killing or violent content] is embodied,” or the “context in which the [killing or violent content] is presented.” But to the extent that the regulation of medium and context are related, the majority’s rule will be unworkable. 

The child who might otherwise play The Last of Us might otherwise play The Path, instead, because the former game has fewer graphic content and the latter is rated M for Mature. And the child who might otherwise enjoy the violent video game The Walking Dead might otherwise have played might instead have played The Secret of Monkey Island, which has no violence or blood and is rated A. The child who might otherwise have played Grand Theft Auto might instead have played The Legend of Zelda, which has no violence or blood and is rated E. And so on. The variety of content that the child may choose from makes it virtually impossible to determine with any precision what the regulation of what the child sees and does not see will look like. The only thing that the majority seems to be able to say with any precision is that the regulation will likely contain some content-based exceptions. But as the San Francisco ordinance demonstrates, those exceptions are exceedingly rare. And when they do appear, they almost always involve content the State has determined is not suitable for children. 

I fear that the majority’s rule will produce a generation of Americans—white, middle-class, suburban—who have no interest in, and no capacity to, engage in the kinds of play the First Amendment protects. And I fear that this Court, which prides itself on its willingness to protect the most vulnerable members of our society from harm, will be complicit in that generation’s continued marginalization. I respectfully dissent.',

\paragraph{Decision:} `deny'

\bigskip
\end{document}